\documentclass[acmtog]{acmart}
\acmSubmissionID{1790}

\usepackage{booktabs} % For formal tables

\usepackage[ruled]{algorithm2e} % For algorithms
\usepackage{multirow}

\SetAlFnt{\small}
\SetAlCapFnt{\small}
\SetAlCapNameFnt{\small}
\SetAlCapHSkip{0pt}

\acmJournal{TOG}

\usepackage{xcolor}

\newcommand{\NICKNAME}{\textsc{ArtiLatent}} % temp "E"ncoder "3D" "G"an inv"E"rsion
\newcommand{\nickname}{\NICKNAME} % temp "E"ncoder "3D" "G"an inv"E"rsion

\newcommand*{\eg}{e.g.\@\xspace}

\copyrightyear{2025}
\acmYear{2025}
\setcopyright{acmlicensed}\acmConference[SA Conference Papers '25]{SIGGRAPH Asia 2025 Conference Papers}{December 15--18, 2025}{Hong Kong, Hong Kong}
\acmBooktitle{SIGGRAPH Asia 2025 Conference Papers (SA Conference Papers '25), December 15--18, 2025, Hong Kong, Hong Kong}
\acmDOI{10.1145/3757377.3763922}
\acmISBN{979-8-4007-2137-3/2025/12}

\begin{document}
\title{\nickname{}: Realistic Articulated 3D Object Generation via Structured Latents
}

% DO NOT ENTER AUTHOR INFORMATION FOR ANONYMOUS TECHNICAL PAPER SUBMISSIONS TO SIGGRAPH 2019!
\author{Honghua Chen}
\affiliation{\institution{S-Lab, Nanyang Technological University}\country{Singapore}}\email{chenhonghuacn@gmail.com}
\author{Yushi Lan}
\affiliation{\institution{S-Lab, Nanyang Technological University}\country{Singapore}}\email{yushi001@e.ntu.edu.sg}
\author{Yongwei Chen}
\affiliation{\institution{S-Lab, Nanyang Technological University}\country{Singapore}}\email{yongwei001@e.ntu.edu.sg}
\author{Xingang Pan}
\affiliation{\institution{S-Lab, Nanyang Technological University}\country{Singapore}}\email{xingang.pan@ntu.edu.sg}

\begin{abstract}
We propose \nickname{}, a generative framework that synthesizes human-made 3D objects with fine-grained geometry, accurate articulation, and realistic appearance. Our approach jointly models part geometry and articulation dynamics by embedding sparse voxel representations and associated articulation properties—including joint type, axis, origin, range, and part category—into a unified latent space via a variational autoencoder. A latent diffusion model is then trained over this space to enable diverse yet physically plausible sampling.
To reconstruct photorealistic 3D shapes, we introduce an articulation-aware Gaussian decoder that accounts for articulation-dependent visibility changes (e.g., revealing the interior of a drawer when opened). By conditioning appearance decoding on articulation state, our method assigns plausible texture features to regions that are typically occluded in static poses, significantly improving visual realism across articulation configurations.
Extensive experiments on furniture-like objects from PartNet-Mobility and ACD datasets demonstrate that ArtiLatent outperforms existing approaches in geometric consistency and appearance fidelity. Our framework provides a scalable solution for articulated 3D object synthesis and manipulation.
Project page: \url{https://chenhonghua.github.io/MyProjects/ArtiLatent/}
\end{abstract}



%
% End generated code
%


\begin{teaserfigure}
  \centering
  \includegraphics[width=\textwidth]{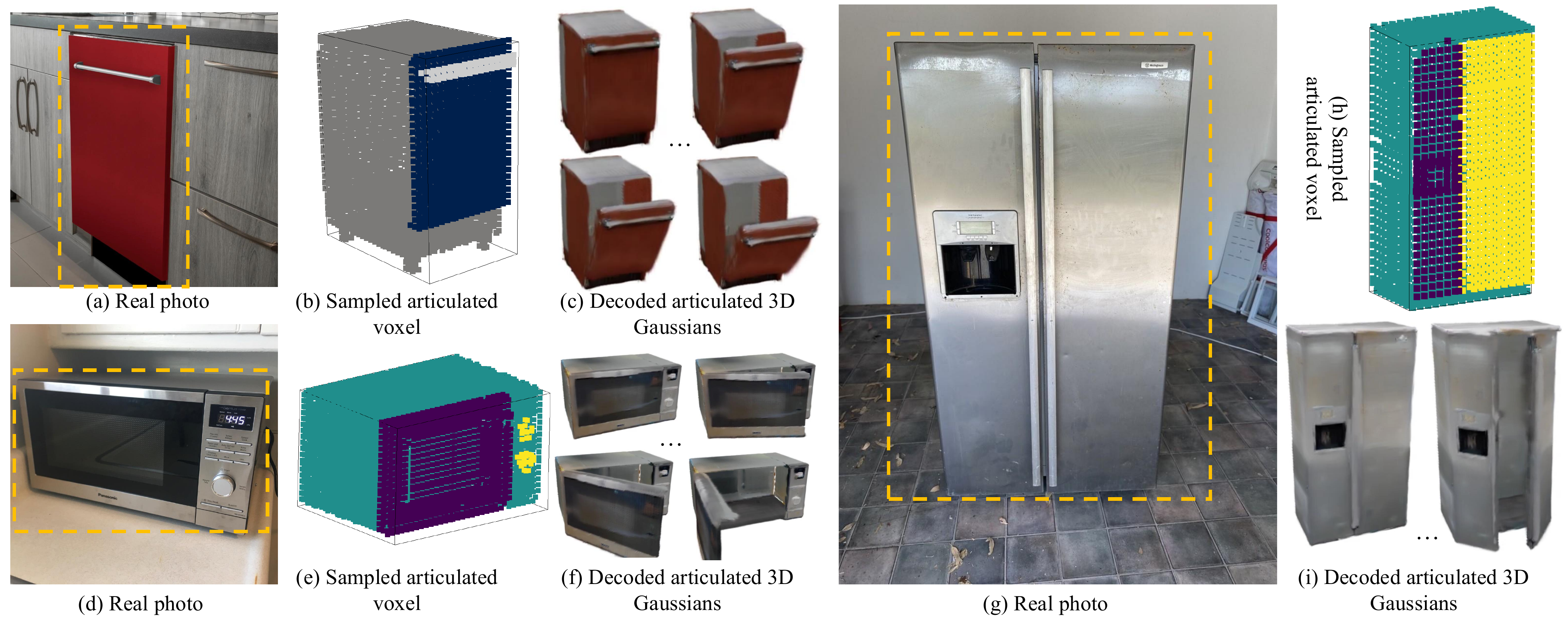}
    \caption{
    Real image conditioned generation of articulated 3D objects.
    Given a real-world image (a, d, g) as input condition, our framework generates articulated 3D objects with realistic geometry, articulation, and appearance. For each example, we first generate an articulation-aware voxel structure (b, e, h), and then decode it into 3D Gaussian splats that support physically plausible part-level motion (c, f, i). The resulting models exhibit high visual fidelity and motion consistency across various object types. Note that we crop out the target object from each scene to serve as the condition image. 
    }
  \label{fig:teaser}
\end{teaserfigure}

\maketitle

\section{Introduction}
Human-made articulated 3D objects comprise multiple semantically meaningful parts connected by joints with constrained motion. They serve as interactive, functional, and physically plausible assets in virtual and physical environments, ranging from everyday items like chairs and drawers to complex industrial tools.
Modeling and generating such objects requires simultaneously capturing three tightly coupled aspects: fine-grained geometry, part-level articulation behavior, and realistic appearance.
This capability is essential to a variety of applications such as high-fidelity simulation, immersive virtual environments, and embodied AI.

Recent progress in 3D generative modeling has yielded impressive results in static object generation, as shown by models such as 
CLAY~\citep{zhang2024clay},
GaussianAnything~\citep{yushigaussiananything}, 3DTopiaXL~\citep{chen20243dtopia}, and TRELLIS~\citep{xiang2024structured}.
However, these methods primarily focus on modeling global geometry and appearance distributions of rigid, non-articulated shapes, but cannot produce physically plausible, part-aware articulation.

To address this, emerging approaches such as NAP~\citep{lei2023nap} and CAGE~\citep{liu2024cage} attempt to jointly model object structure and articulation properties. These methods represent geometry with coarse bounding boxes and generate final shapes using implicit fields or retrieval-based part assembly. However, such approaches often lead to inconsistent geometry and suboptimal inter-part alignment. SINGAPO~\citep{liu2024singapo} introduces image conditioning into this framework but retains the limitations of retrieval-based pipelines. MeshArt~\citep{gao2024meshart} takes a step toward higher-fidelity modeling using triangle meshes, yet it focuses solely on geometry and lacks texture modeling, limiting its applicability in photorealistic or interactive scenarios.

Meanwhile, other methods such as PARIS~\citep{liu2023paris} and ArticulatedGS~\citep{guo2025articulatedgs,liu2025artgs} focus on reconstructing articulated objects from multi-state inputs. While effective at recovering geometry and motion from paired observations, they rely on pre-captured start and end states. More recent approaches, including Articulate AnyMesh~\citep{qiu2025articulate} and ATOP~\citep{vora2025articulate}, attempt to infer articulation given a static mesh. However, all of these methods are reconstructive in nature and do not support generative modeling or conditional synthesis.

In this work, our goal is to develop a generative framework capable of synthesizing articulated 3D objects with fine-grained geometry, appearance, and part-level articulation properties. Moreover, we aim to support conditional generation from real-world images to broaden applicability across practical scenarios.
To achieve this, we make three key designs:

First, previous methods~\citep{lei2023nap,liu2024cage,liu2024singapo} model each part independently, typically assuming a fixed upper bound on part count and relying on retrieval-based assembly. This leads to limited scalability and inaccurate or inconsistent part arrangement. In contrast, we adopt a structured global representation based on sparse voxels. This design is motivated by recent advances in static 3D object generation, where methods such as TRELLIS~\citep{xiang2024structured} and GaussianAnything~\citep{yushigaussiananything} demonstrate that adopting structured 3D as the latent space (\eg, sparse voxels or point clouds), when paired with 3D Gaussian decoders, can effectively capture high-quality, globally coherent geometry with natural inter-part continuity. To fully leverage learned priors of generating photorealistic rigid 3D objects, we use sparse voxels as our coarse geometric representation.

Second, for modeling articulation, we observe that object geometry, part semantics, and articulation properties are intrinsically intertwined—part shape and function often imply specific joint types and constraints (\eg, drawers tend to translate, while doors typically rotate). To capture these correlations, we jointly embed sparse voxels, part category labels, and associated articulation attributes (\eg, joint type, axis, origin, and range) into a unified latent space via a variational autoencoder (VAE). By attaching local joint parameters and semantic tags to each voxel, 
the latent encodes shape, semantics, and articulation in a consistent and integrated manner, facilitating the learning of their joint distribution via a diffusion model.
With this design, the diffusion model can operate in the latent space and generate physically plausible articulated structures.
This design allows the model to effectively capture the underlying correlations and generate physically plausible articulated structures.

Third, for appearance generation, we leverage the structured latent diffusion model of TRELLIS by sampling latent codes for all voxels and decoding them into textured 3D Gaussians. 
However, TRELLIS's pretrained model is unaware of visibility changes caused by articulation, for instance, newly exposed surfaces (\eg, inside a drawer) often exhibit unrealistic textures when articulation alters visibility. This is primarily because occluded regions receive little or no supervision during 3D VAE pretraining, leading to uninformative latent features. To alleviate this issue, we propose an articulation-aware fine-tuning strategy that supervises the 3D VAE autoencoding using rendered images of ``transformed'' 3D Gaussians across different articulation states. This enables the latent codes to adapt to articulation-aware appearance variations, resulting in more realistic and consistent texture synthesis.

In summary, we present \nickname{}, a diffusion-based framework that jointly models shape, articulation, and appearance to generate high-fidelity, human-made articulated 3D objects. Extensive evaluations on two articulated object benchmarks show that ArtiLatent consistently outperforms existing methods in motion controllability, geometric coherence, and appearance fidelity. Our approach also enables generating a complete articulated 3D object from a single real-world image, as illustrated in Fig.~\ref{fig:teaser}, while faithfully preserving the appearance in the input. With its enhanced visual fidelity and articulation modeling, our method represents a significant step toward realistic and interactive 3D environments, laying the foundation for downstream applications such as embodied AI and digital twin construction.
\label{sec:intro}

\section{Related Work}
\subsection{3D/4D Object Generation}
3D generative models, especially 3D latent diffusion models, have recently shown 
remarkable capabilities in synthesizing high-quality, efficient, and scalable 3D objects. 
~\citep{zhang20233dshape2vecset,zhang2024clay,xiang2024structured,yushigaussiananything,chen20243dtopia,zhao2025hunyuan3d,li2025triposg,lan2024ln3diff,chen2025primx}. 
However, they primarily focus on modeling geometry and textures of static, non-articulated 3D objects and fail to capture part-level structure and motion.

Beyond static generation, 4D object modeling focuses on capturing temporal dynamics such as object motion and deformation over time~\citep{ren2023dreamgaussian4d,gao2024gaussianflow,zeng2024stag4d}. These approaches typically model continuous motion via deformation fields~\citep{park2021nerfies,lan2022ddf_ijcv} or time-varying geometry.
However, they do not explicitly model discrete, joint-based articulation or encode semantic part structure. 
Moreover, they are not designed for human-made objects composed of rigid parts connected via articulated joints, where motion follows structural and kinematic constraints.
In contrast, our work targets the generation of articulated 3D objects with detailed geometry and explicitly controllable joint-level motion. 

\subsection{Structured Data Generation}
Our task is also related to structured 3D data generation~\citep{chaudhuri2020learning}, which focuses on synthesizing shapes composed of semantically meaningful and geometrically coherent parts. Earlier works tackled this problem using voxel grids with semantic labels~\citep{wang2018global,li2020learning,wu2020pq}, latent space reasoning with structural priors such as symmetry and support~\citep{wu2019sagnet}, or explicit part hierarchies modeled through tree-based architectures~\citep{li2017grass,mo2019structurenet,gao2019sdm}.
Another line of research investigates 3D assembly, where complex shapes are composed by arranging primitives~\citep{gadelha2020learning,paschalidou2021atiss,jones2020shapeassembly,xu2024brepgen}, joints~\citep{willis2022joinable,li2024category} or semantic parts~\citep{zhan2020generative,li2020learning,narayan2022rgl,xu2023unsupervised,koo2023salad}. Structured generation has also been extended to scene composition~\citep{wang2021sceneformer,wei2023lego,tang2024diffuscene} and architectural layout synthesis~\citep{nauata2020house,nauata2021housegan,shabani2023housediffusion,tang2023graph}, where spatial and relational constraints are explicitly encoded.

Articulated objects represent a special class of structured data, in which part geometry and motion are inherently coupled. Generating such objects requires not only part coherence but also consistency in joint behavior and motion feasibility~\citep{liu2024survey}. 
Our method leverages structured global sparse voxels and explicit motion attributes, enabling controllable and geometry-consistent articulation generation without retrieval-based post-assembly.

\subsection{Articulated Object Modeling and Generation}
Articulated object modeling has been extensively studied in the contexts of reconstruction and motion analysis. Early methods such as Shape2Motion~\citep{wang2019shape2motion}, ScrewNet~\citep{jain2021screwnet}, PARIS~\citep{liu2023paris}, and ArticulatedGS~\citep{liu2025artgs,guo2025articulatedgs} focus on part segmentation and joint parameter estimation from multi-view or multi-state observations.
Later, when only static observations are available, DRAWER~\citep{xia2025drawer} converts a single-view video of a static scene into an interactive and actionable virtual environment.
Building on incomplete geometric inputs, PhysPart~\citep{luo2024physpart} imposes physical constraints through stability and mobility losses to guide the generation of animatable parts.
More recently, ATOP~\citep{vora2025articulate} introduced a video-conditioned pipeline that animates existing 3D assets through motion transfer. However, it does not support object-level generation from scratch, and its category-specific design limits its ability to generalize to unseen object categories.
To address these limitations, DreamArt~\citep{lu2025DreamArt} learns a more generalizable motion prior by leveraging a more intuitive and readily available control signal, namely the movable part mask.

Generative approaches such as NAP~\citep{lei2023nap}, CAGE~\citep{liu2024cage}, MeshArt~\citep{gao2024meshart}, and ArtFormer~\citep{su2025artformer} focus on synthesizing articulated objects with controllable structures. CAGE leverages part-graph constraints for joint control, while MeshArt and ArtFormer improve geometry realism through transformer-based mesh generation or SDF-based geometry decoder.
However, these methods either ignore appearance modeling or rely on part retrieval and assembly, limiting flexibility and expressiveness. Infinite Mobility~\citep{lian2025infinite} adopts a procedural pipeline for generating large-scale articulated objects, but still requires mesh retrieval and post-refinement to obtain usable outputs. 
In contrast, our approach unifies geometry and motion modeling within a shared latent space and introduces an articulation-aware appearance decoder. This design enables direct generation of photorealistic, structurally consistent articulated 3D objects, supporting fine-grained motion control and diverse conditional generation.
\label{sec:related}

% =============================================== Preliminaries ==================================================

\begin{figure*}[t]
    \centering
    \includegraphics[width=\linewidth]{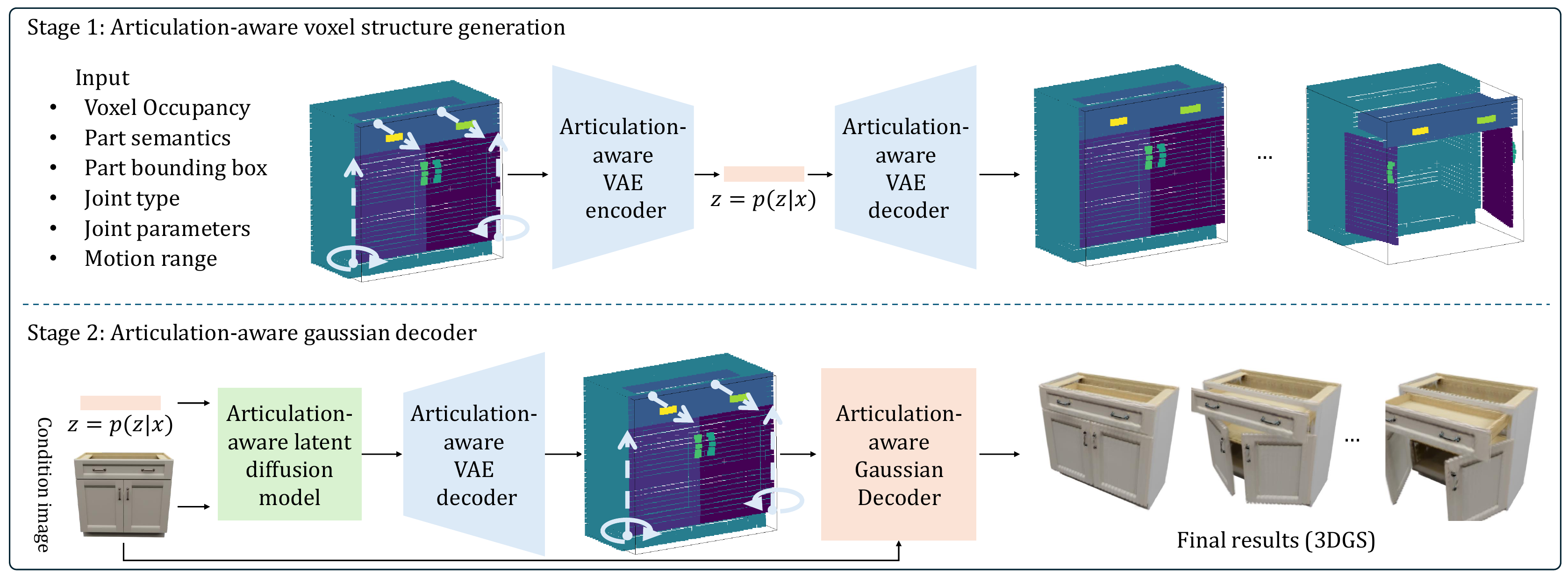}
    \caption{Method overview. Given voxel-level articulation-aware inputs (occupancy, semantics, joint types, bounding boxes, joint parameters, and motion ranges), we encode them into a latent representation using an articulation-aware VAE. A conditional diffusion model samples articulation-aware latent codes under user-specified conditions (\eg, image), which are then decoded into an animatable voxel structure. The final appearance is generated using an articulation-aware Gaussian decoder, producing high-fidelity 3D Gaussian splats with consistent geometry and appearance across motion states.}
    \label{fig:overview}
\end{figure*}

\section{Preliminaries}
\label{sec:Preliminaries}

Our method builds upon TRELLIS~\cite{xiang2024structured}, a recent framework for high-quality 3D generation. TRELLIS establishes a scalable encoding scheme. Each 3D asset is first converted into voxelized features, where active voxels aggregate local geometry and appearance information from multi-view renderings processed by a pretrained DINOv2 encoder. This feature grid, aligned with the latent resolution (e.g., $64^3$), captures both coarse structural priors and fine visual details. 
A transformer-based sparse VAE then encodes these voxelized features into structured latent codes and decodes them back into various 3D formats using modality-specific decoders, such as the Gaussian splat decoder $\mathcal{D}_{\mathrm{GS}}^{\mathrm{Tre}}$~\cite{kerbl20233d}. 
Its generation pipeline involves two stages: (1) a rectified flow model~\cite{esser2024scaling} predicts a dense occupancy grid, which is converted into a sparse voxel structure; (2) a sparse rectified flow transformer then generates the structured latent conditioned on this geometry. 

In our approach, we leverage TRELLIS’s latent diffusion model $\mathcal{G}^{\mathrm{Tre}}$ to sample voxel-wise latents, which are then decoded into high-fidelity 3D Gaussians by $\mathcal{D}_{\mathrm{GS}}^{\mathrm{Tre}}$ to generate photorealistic 3D objects. Importantly, we leverage TRELLIS’s pretrained weights, which contribute to improved training stability and generalization.

\section{Method}
\label{sec:method}
In this section, we introduce \nickname{}, a generative framework that synthesizes articulated 3D objects with fine-grained geometry, physically plausible part-level motion, and realistic appearance. Our method consists of two key stages (see Fig.~\ref{fig:overview}): (1) generating an articulation-aware sparse voxel representation that encodes both geometry and motion; and (2) reconstructing photorealistic articulated objects via an articulation-aware Gaussian decoder.

\subsection{Articulation-aware voxel structure generation}

\subsubsection{Articulated voxel representation}
We represent an articulated object as a sparse 3D voxel field, where each voxel $v_i$ corresponds to a localized volumetric region and is associated with rich semantic, geometric, and motion-related attributes. Specifically, for each active voxel, we attach the following information:
\begin{itemize}
    \item \textbf{Occupancy}: Following~\citet{xiang2024structured}, we convert the sparse voxel set into a dense binary occupancy grid $\mathbf{O} \in \{0, 1\}^{N \times N \times N}$, where $\mathbf{O}(x, y, z) = 1$ if the corresponding voxel intersects the object, and $0$ otherwise. We set $N = 64$.

    \item \textbf{Part semantics}: A categorical label $l_i \in$ \{base, drawer, door, handle, knob, tray, shelf, wheel\}, represented via one-hot encoding.
    
    \item \textbf{Part bounding box}: A bounding box $b_i \in \mathbb{R}^6$ describing the 3D center and size of the part associated with voxel $v_i$.

    \item \textbf{Joint type}: A discrete label $j_i \in$ \{fixed, revolute, prismatic, continuous, screw\}, also one-hot encoded.

    \item \textbf{Joint parameters}: A joint axis $a_i \in \mathbb{R}^3$ and origin $o_i \in \mathbb{R}^3$, specifying the direction and position of the joint's motion.

    \item \textbf{Motion range}: A joint limit $r_i \in \mathbb{R}^2$, encoding the allowed angular or translational motion range.

\end{itemize}
All voxel-level attributes are normalized in a canonical coordinate space, where each object is centered and consistently oriented, following~\citet{liu2024cage,liu2024singapo}. Notably, all voxels belonging to the same part instance share identical semantic and articulation attributes.

\subsubsection{Articulation-aware latent compression}
We employ a VAE $\{\mathcal{E}^{\mathrm{Arti}}, \mathcal{D}^{\mathrm{Arti}}$\} with 3D convolutional blocks to encode the articulation-aware voxel representation into a compact latent space. The encoder $\mathcal{E}^{\mathrm{Arti}}$ takes as input a dense volumetric tensor of shape $[C_{\text{in}}, 64, 64, 64]$, where $C_{\text{in}} = 35$. This includes one channel for the binary occupancy grid $\mathbf{O}$ and 34 channels encoding voxel-level articulation attributes, such as one-hot part labels, one-hot joint types, joint axes and origins, motion ranges, and bounding box parameters.
The encoder outputs a latent volume of shape $[2C_z, 16, 16, 16]$ with $C_z = 8$, representing the mean and log-variance used to sample the latent variable $z \in \mathbb{R}^{C_z \times 16 \times 16 \times 16}$ via the reparameterization trick. The decoder $\mathcal{D}^{\mathrm{Arti}}$ mirrors the encoder architecture with upsampling blocks, and reconstructs a volumetric output of shape $[C_{\text{out}}, 64, 64, 64]$, predicting per-voxel occupancy, semantics, and articulation attributes.

\paragraph{Training Loss.} 
We train the VAE using a combination of reconstruction and regularization objectives. A KL-regularized loss is applied to enforce a continuous and generative latent space. For voxel-wise reconstruction, we design attribute-specific loss terms:
\begin{itemize}
    \item \textbf{Occupancy classification}: We adopt the Dice loss~\citep{milletari2016v} to mitigate the severe class imbalance between occupied and unoccupied regions. 
    For ground-truth occupancy labels $y_i \in \{0,1\}$ and predictions $\hat{y}_i \in [0,1]$,  
    \begin{equation}
    \mathcal{L}_{\mathrm{occ}} = 1 - 
    \frac{2 \sum_{j=1}^{M} y_j \hat{y}_j}{\sum_{j=1}^{M} y_j + \sum_{j=1}^{M} \hat{y}_j + \epsilon},
    \end{equation}
    where $M=N^3$ denotes the total number of voxels and $\epsilon$ is a small constant for numerical stability.
    
    \item \textbf{Part semantic type classification}: We apply cross-entropy loss to supervise the predictions of part semantic labels and joint types.
    With one-hot labels $l_{i,c}^{\mathrm{sem}} \in \{0,1\}$ and predicted probabilities $\hat{p}_{i,c}^{\mathrm{sem}}$,  
    \begin{equation}
    \mathcal{L}_{\mathrm{sem}} = - \frac{1}{M'} \sum_{i=1}^{M'} \sum_{c=1}^{C_{\mathrm{sem}}} 
    l_{i,c}^{\mathrm{sem}} \log \hat{p}_{i,c}^{\mathrm{sem}},
    \end{equation}
    where $M'$ is the total number of active voxels and $C_{\mathrm{sem}}$ is the number of semantic part categories, 

    \item \textbf{Joint type classification}:  
    With one-hot labels $j_{i,c}^{\mathrm{joint}} \in \{0,1\}$ and predicted probabilities $\hat{p}_{i,c}^{\mathrm{joint}}$,  
    \begin{equation}
    \mathcal{L}_{\mathrm{joint}} = - \frac{1}{M'} \sum_{i=1}^{M'} \sum_{c=1}^{C_{\mathrm{joint}}} 
    j_{i,c}^{\mathrm{joint}} \log \hat{p}_{i,c}^{\mathrm{joint}},
    \end{equation}
    where $C_{\mathrm{joint}}$ is the number of joint type categories.
    
    \item \textbf{Articulation parameter regression}: 
    For continuous attributes (joint axis vectors $\hat{a}_i$, origins $\hat{o}_i$, motion ranges $\hat{r}_i$, and bounding box parameters $\hat{b}_i$) with ground truth values $(a_i, o_i, r_i, b_i)$,  
    \begin{equation}
    \mathcal{L}_{\mathrm{bbox}} = 
    \frac{1}{M'}\sum_{i=1}M^{'} \Big( 
    \| a_i - \hat{a}_i \|_2^2 +
    \| o_i - \hat{o}_i \|_2^2 +
    \| r_i - \hat{r}_i \|_2^2 +
    \| b_i - \hat{b}_i \|_2^2
    \Big).
    \end{equation}

\end{itemize}
The total loss is a weighted sum of all components:
\begin{equation}{
 \mathcal{L}_{\mathrm{total}} = \alpha_{\mathrm{kl}}\mathcal{L}_{\mathrm{KL}} + \mathcal{L}_{\mathrm{occ}} + \mathcal{L}_{\mathrm{sem}} + \mathcal{L}_{\mathrm{joint}} +  \mathcal{L}_{\mathrm{bbox}}.
\label{eq:vae}}
\end{equation}

This training objective ensures accurate reconstruction of voxel-level semantics and motion attributes, while maintaining a smooth and generative latent representation.

\subsubsection{Articulation-aware latent generation}
After training the VAE model, we obtain a latent dataset consisting of $D$ samples, where each sample is a pair of a latent code and a corresponding condition vector: $\{(z_i, c_i)\}_{i=1}^{D}$. Here, $c_i$ encodes external conditioning information (\eg, image embedding or text prompt).

To enable conditional generative modeling, we train a flow matching network~\citep{lipmanflow}, $\mathcal{G}^{\mathrm{Arti}}$, to learn a diffusion prior to the latent space. Following TRELLIS, we adopt a transformer-based denoising backbone that processes serialized latent grids with 3D positional encodings. For details of the architecture, we refer readers to~\citet{xiang2024structured}.
We also incorporate classifier-free guidance to flexibly inject various types of conditioning, including category labels or visual embeddings from DINOv2~\citep{oquab2023dinov2}. This enables the model to generate diverse and physically plausible latent codes under different user-specified prompts, which are subsequently decoded by the VAE decoder into high-fidelity articulated 3D voxelized objects.

\begin{figure}[t]
  \centering
  \includegraphics[width=\linewidth]{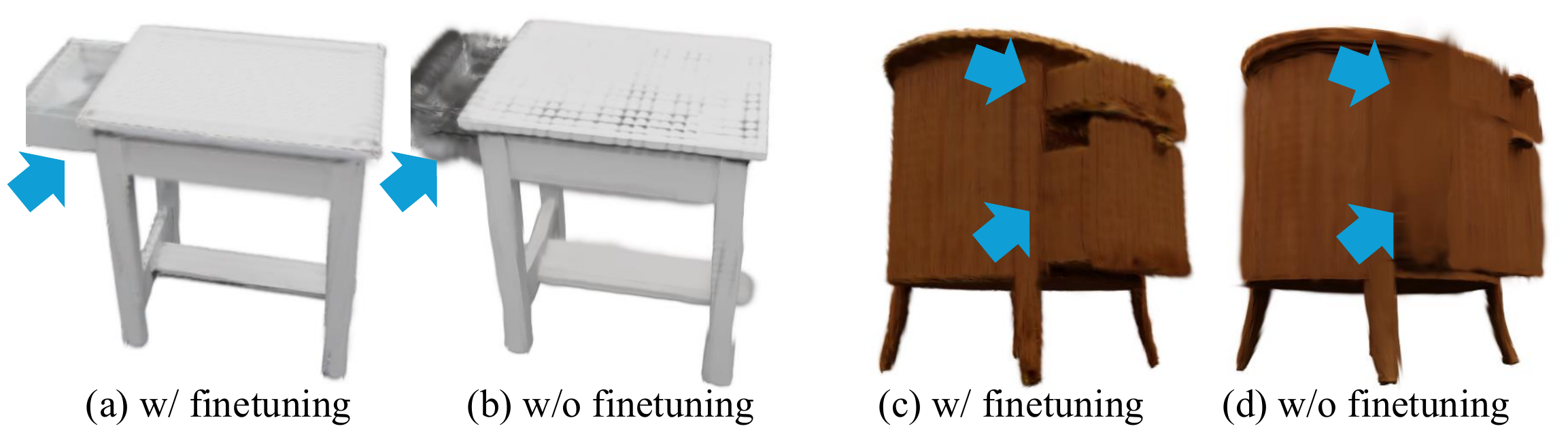}
    \caption{
    Effect of articulation-aware fine-tuning on appearance quality. 
    We compare the results with (a, c) and without (b, d) articulation-aware fine-tuning on two different object types. Without fine-tuning, the generated textures exhibit noticeable artifacts, such as distortion, color bleeding, and loss of structure in articulated regions (see blue arrows). In contrast, our fine-tuned model produces sharper, more consistent, and plausible textures, especially around seams and occluded parts revealed by motion.
    }
  \label{fig:stage2_results}
\end{figure}

\subsection{Articulation-aware Gaussian decoder}
With the ability to freely sample articulated voxel structures, our goal is to reconstruct high-fidelity 3D Gaussian splats. To this end, we leverage the structured latent diffusion model $\mathcal{G}^{\mathrm{Tre}}$ and the Gaussian decoder $\mathcal{D}_{\mathrm{GS}}$ from TRELLIS~\citep{xiang2024structured}. Specifically, for each voxel, we sample a latent feature $z^{\mathrm{Tre}}$ using $\mathcal{G}^{\mathrm{Tre}}$ and decode it into a 3D Gaussian representation via $\mathcal{D}_{\mathrm{GS}}$.

\begin{figure}[t]
    \centering
    \includegraphics[width=\linewidth]{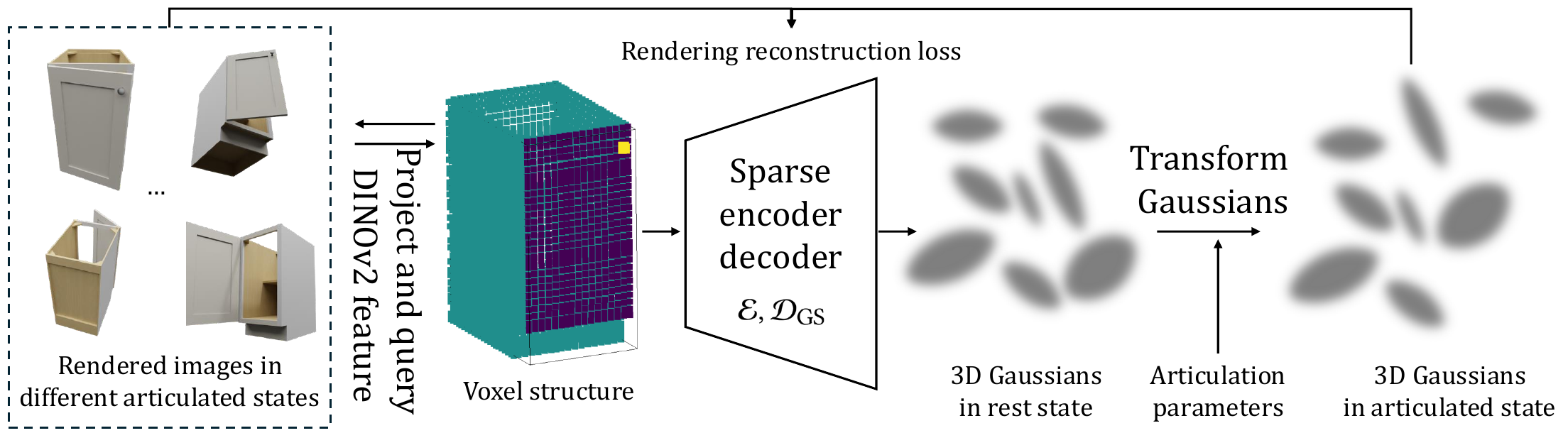}
    \caption{Articulation-aware Gaussian decoder. We render the generated 3D Gaussians under multiple articulated states and use the corresponding images to supervise the encoder-decoder pair ($\mathcal{E}, \mathcal{D}_{\mathrm{GS}}$). For each voxel, we extract DINOv2 features across states and views, and use them as the initial feature. Decoded Gaussians are transformed according to articulation parameters, enabling the model to learn articulation-aware appearance variations via reconstruction loss.}
    \label{fig:stage2}
\end{figure}

However, directly applying $\mathcal{D}_{\mathrm{GS}}$ to articulated objects often results in suboptimal appearance, particularly in regions that are occluded in the closed state but become visible after articulation. As illustrated in Fig.~\ref{fig:stage2_results}, inner surfaces of a drawer may exhibit noisy or unrealistic textures once opened. This issue arises because the original sparse VAE $\{\mathcal{E}, \mathcal{D}_{\mathrm{GS}}\}$ was trained on static objects and lacks exposure to diverse articulation states during training.
To overcome this limitation, we introduce an articulation-aware fine-tuning strategy, as shown in Fig.~\ref{fig:stage2}. We render the same object under multiple articulated poses and use the corresponding 2D views as supervision signals. This allows us to adapt $\{\mathcal{E}, \mathcal{D}_{\mathrm{GS}}, \mathcal{G}^{\mathrm{Tre}}\}$ to become aware of articulation-dependent visibility changes. As a result, the latent $z^{\mathrm{Tre}}$ becomes more informative with respect to articulation-aware geometry exposure, enabling more realistic and consistent texture synthesis across different articulation states.

\subsubsection{Multiple-state data curation}
We curate a multi-state dataset by sampling articulated objects across their full articulation ranges. For each object, we uniformly sample $k$ articulation states and render $n$ views per state, ensuring comprehensive visibility coverage.
For each state, we articulate the voxelized object and extract per-view visual features using DINOv2. For every active voxel, we aggregate its features across all views and articulation states by averaging them. The resulting averaged feature is then assigned to the corresponding voxel in the rest (closed) state, which we use for sampling $z^{\mathrm{Tre}}$. This choice is due to the fact that the conditional image provided during generation typically depicts the object in its rest configuration.

\subsubsection{Articulation-aware fine-tuning}
Once the multi-state training data is prepared, we fine-tune $\{\mathcal{E}, \mathcal{D}_{\mathrm{GS}}, \mathcal{G}^{\mathrm{Tre}}\}$. When training $\{\mathcal{E}, \mathcal{D}_{\mathrm{GS}}\}$, we articulate the decoded Gaussians $\{g_i\}$ and supervise the rendered images with the corresponding ground-truth views.

To perform articulation on $\{g_i\}$, we first establish a mapping between voxels and their corresponding Gaussian points. In TRELLIS, each active voxel is decoded into $32$ Gaussians within its local neighborhood, and the outputs are sequentially ordered. This allows us to retrieve the set of Gaussians associated with a given voxel based on its index.
Using the articulation parameters attached to each voxel, we apply spatial transformations—such as translations or rotations—to the corresponding Gaussians to simulate their articulated state. The transformed Gaussians are then rendered from specific camera views and supervised using reconstruction losses against the corresponding ground-truth images  $I_{gt}^{s}$:
\begin{equation}
\mathcal{L}_{\mathrm{GS}} = \mathcal{L}_{\mathrm{recon}} + \lambda \mathcal{L}_{\mathrm{reg}}, \quad
\mathcal{L}_{\mathrm{recon}} = M\left(I_{gt}^{s}, f_{\mathrm{render}}\left(\{T_i^s(g_i)\}\right)\right),
\label{eq:deform_gs}
\end{equation}
where $M(\cdot)$ is any image reconstruction metric (e.g., $\ell_1$, $\ell_2$, or perceptual loss) computed between the rendered image and the ground-truth image, $T_i^s(\cdot)$ denotes the transformation applied at articulation state $s$ for Gaussian point $g_i$, and $\mathcal{L}_{\mathrm{reg}}$ is a regularization term applied to the predicted Gaussians.

This supervision fine-tunes $\{\mathcal{E}, \mathcal{D}_{\mathrm{GS}}\}$ to capture articulation-aware appearance variations. Once adapted, we re-encode the data to obtain updated latent representations and further tine-tune $\mathcal{G}^{\mathrm{Tre}}$ to enable articulation-aware latent sampling.

\subsection{Inference}
During inference, given a specific user-defined condition (\eg, image), we first sample a latent code $z$ from $\mathcal{G}^{\mathrm{Arti}}$ and decode it into an articulated voxel structure.
Since voxels belonging to the same semantic part are expected to share consistent articulation behavior, we segment object parts based on the predicted semantic labels and bounding box information. Specifically, we first perform a coarse segmentation using the predicted part semantics. However, voxels with identical semantic labels may belong to adjacent but distinct parts. To address this, we further apply a DBSCAN clustering step using bounding box attributes (e.g., centers and sizes), which allows us to separate different parts that share the same semantic category but exhibit different spatial properties. For each segmented part, we then aggregate per-voxel articulation parameters by averaging them within the segment, and assign the aggregated values back to all voxels in that part. This ensures coherent articulation behavior and physically plausible motion at the part level.
Conditioned on the input and the generated voxel structure, we then use $\mathcal{G}^{\mathrm{Tre}}$ to assign latent features to each voxel. These structured features are decoded by $\mathcal{D}_{\mathrm{GS}}$ into 3D Gaussian splats, producing photorealistic reconstructions that capture both exterior and interior surfaces. Importantly, the resulting Gaussians support smooth and realistic articulation by construction.
Note that at both training and inference time, our model takes a single RGB image, depicting the object in rest state from a near-frontal view. The rest-state voxel is used for the 3D Gaussian generation to ensure appearance-geometry alignment. The output is a 3D Gaussian with articulation parameters, enabling articulated motion by joint-driven transformations without re-running inference.

\section{Experiments}

\subsection{Implementation}
For articulation-aware VAE training, we train ${\mathcal{E}^{\mathrm{Arti}}, \mathcal{D}^{\mathrm{Arti}}}$ from scratch using 4×A6000 GPUs for 1 day until convergence. The KL divergence loss term is weighted by $\alpha_{\mathrm{kl}}=0.001$, and the reconstruction objectives are each assigned a weight of 1.
The articulation diffusion model $\mathcal{G}^{\mathrm{Arti}}$ is trained under the same hardware setup for 1 day, initialized from the structure diffusion model pretrained in TRELLIS. For fine-tuning ${\mathcal{E}, \mathcal{D}_{\mathrm{GS}}, \mathcal{G}^{\mathrm{Tre}}}$, we again use 4×A6000 GPUs over 2 days. Note that during articulation and supervision, we uniformly sample $k=8$ articulation states, each rendered from $n=48$ camera views.
All models are optimized with the Adam optimizer, using a learning rate of $1 \times 10^{-4}$ and a batch size of 4 per GPU. During inference, we set the classifier-free guidance (CFG) strength to 3 and the number of sampling steps to 50. 

\subsection{Dataset}
We conduct our experiments on a subset of the PartNet-Mobility dataset~\citep{xiang2020sapien}, focusing on seven common categories: {\small\texttt{Storage}}, {\small\texttt{Table}}, {\small\texttt{Refrigerator}}, {\small\texttt{Dishwasher}}, {\small\texttt{Oven}}, {\small\texttt{Washer}}, and {\small\texttt{Microwave}}. The dataset is preprocessed following~\citep{liu2024singapo}, resulting in 3,063 articulated objects for training. For evaluation, we use 77 held-out instances, each paired with two randomly rendered views to simulate conditional inputs.
To assess generalization beyond the training distribution, we  also evaluate our model in a zero-shot setting using 135 unseen objects from the ACD dataset~\citep{iliash2024s2o}. Additional preprocessing and dataset construction details are consistent with prior work~\citep{liu2024singapo}.

\subsection{Baselines and evaluation metrics}
Since our method supports the conditional generation of articulated 3D objects, we compare against representative baselines under the image-conditioned setting. Specifically, we include SINGAPO~\citep{liu2024singapo}, a state-of-the-art controllable generation model that takes a single image as input. As we use the same training and test datasets, we directly report the official results from their paper.
For broader comparison, we also include NAP-ICA, the image-conditioned variant of NAP, as introduced in~\citep{liu2024singapo}.

\noindent\textbf{Evaluation Metrics.}  
We adopt several metrics to evaluate geometric accuracy and visual realism of articulated 3D object generation. 

\begin{itemize}
    \item $d_{\text{CD}} \downarrow$: Chamfer Distance (CD) between sampled surface points across articulated states, measuring geometric alignment. More specifically, RS-$d_{\text{CD}}$ refers to the CD value computed in the rest state, while AS-$d_{\text{CD}}$ denotes the distance measured after articulation. 

    \item FID $\downarrow$: Fréchet Inception Distance computed between rendered images of the generated shapes (Gaussian splats or retrieved meshes) and those of the ground-truth meshes, assessing perceptual fidelity.
\end{itemize}
Note that during evaluation, we render two views of each object in its rest state and randomly select one as the input. Our method performs a single forward pass to generate a 3D Gaussian representation in this rest configuration. To evaluate articulation behavior, we then apply joint-based transformations to the generated 3D Gaussian to simulate five target articulation states. All metrics are computed over these five transformed outputs.

\subsection{Results}
\paragraph{Visual Comparisons}
Fig.~\ref{fig:visual_compasion} and Fig.~\ref{fig:additional_results} present qualitative comparisons across various categories from the PartNet-Mobility and ACD datasets. Compared to SINGAPO, our method generates more accurate part geometry and more realistic textures. Notably, it better captures motion-aware articulation behaviors—such as drawer translations and washer door rotations—and preserves fine-grained appearance details in both exterior surfaces and newly exposed interior regions. 
In contrast, SINGAPO, which relies on part retrieval and mesh assembly, is prone to retrieval mismatches. For example, in the last row of Fig.~\ref{fig:visual_compasion}, it fails to retrieve a correct door geometry for the washing machine, resulting in a shape that does not match the underlying articulation structure. This highlights the advantage of our generative approach in maintaining part-motion consistency and global structural coherence.

\paragraph{Quantitative Comparisons}
Table~\ref{tab:quant_pm_acd} presents the quantitative results on the evaluated datasets under the image-conditioned setting. We assess both geometric accuracy and perceptual quality using RS-$d_{\text{CD}}$, AS-$d_{\text{CD}}$, and FID.
We first compare our method with TRELLIS in the rest state on the PartNet-Mobility test set. TRELLIS achieves a CD of 0.0051 and an FID of 153.45, while our method obtains a CD of 0.0063 and an FID of 137.18, indicating comparable performance in static settings. However, as TRELLIS does not support articulated modeling, it cannot be evaluated under articulation-aware metrics.
For methods that involve articulation modeling, our method consistently outperforms all baselines across both datasets. Specifically, we achieve the lowest CD values in both the rest and articulated states, demonstrating superior geometric reconstruction and articulation consistency. Furthermore, our method yields the lowest FID scores, indicating more realistic visual quality compared to retrieval-based approaches.

\begin{table}[t]
    \caption{
    Quantitative comparison of reconstruction and perceptual quality on the PartNet-Mobility and ACD test sets under the single-image input setting. All methods generate one articulated object per input. RS-$d_{\text{CD}}$ refers to the Chamfer Distance computed in the rest state, while AS-$d_{\text{CD}}$ denotes the distance measured after articulation. Lower is better for all metrics. 
    }
    \label{tab:quant_pm_acd}
    \centering
    \small % or \footnotesize
    \resizebox{0.5\textwidth}{!}{
    \begin{tabular}{@{}lccccccccc@{}}
    \toprule
    \multirow{2}{*}{Method} & \multicolumn{3}{c}{PartNet-Mobility} & \multicolumn{3}{c}{ACD Test Set} \\
    \cmidrule(l){2-4} \cmidrule(l){5-7}
    & RS-$d_{\text{CD}}$ $\downarrow$ & AS-$d_{\text{CD}}$ $\downarrow$ & FID $\downarrow$ 
    & RS-$d_{\text{CD}}$ $\downarrow$ & AS-$d_{\text{CD}}$ $\downarrow$ & FID $\downarrow$ \\
    \midrule
    TRELLIS   & \textbf{0.0051} & -- & 153.45 & -- & -- & -- \\
    URDFormer & 0.5502 & 0.8374 & -- & 0.7198 & 0.8995 & -- \\
    NAP-ICA   & 0.0173 & 0.0914 & -- & 0.1110 & 0.1887 & -- \\
    SINGAPO   & 0.0168 & 0.0905 & 175.85 & 0.1011 & 0.1679 & 201.60 \\
    Ours      & 0.0063 & \textbf{0.0043} & \textbf{137.18}  
              & \textbf{0.0690} & \textbf{0.0751} & \textbf{128.34} \\
    \bottomrule
    \end{tabular}}
\end{table}

\begin{table}[t]
    \caption{
    Ablation study. Incorporating the articulation-aware fine-tuning strategy enables our model to generate more realistic objects.
    }
    \label{tab:quant_abla}
    \centering
    \small % or \footnotesize
    \resizebox{0.5\textwidth}{!}{
    \begin{tabular}{@{}lcccccc@{}}
    \toprule
    \multirow{2}{*}{Method} & \multicolumn{3}{c}{PartNet-Mobility}  \\
    \cmidrule(l){2-4}
    & RS-$d_{\text{CD}}$ $\downarrow$ & AS-$d_{\text{CD}}$ $\downarrow$ & FID $\downarrow$ 
    \\
    \midrule
    w/o Articulation-aware fine-tuning & 0.0076 & 0.0051 & 156.02 \\
    Ours      & \textbf{0.0063} & \textbf{0.0043} & \textbf{137.18} 
              \\
    \bottomrule
    \end{tabular}}
\end{table}

\paragraph{Quantitative evaluation of the predictions of part semantics and articulation parameters}
We evaluated occupancy classification and joint parameter accuracy on the PartNet-Mobility test set, comparing with SINGAPO, as shown in Table~\ref{tab:vae_comparison}. Our method achieves competitive results. 
Additionally, we computed the standard deviation ($std$) of predicted articulation parameters across voxels within each part and observed that the intra-part variance is generally low (see Table~\ref{tab:std}, computed on the PartNet-Mobility test set and averaged over all parts). This supports the spatial consistency of voxel-wise predictions and justifies our averaging strategy.

\begin{table}[ht]
\centering
\caption{Comparison of predictions of part semantics and articulation parameters between SINGAPO and our method.}

\begin{tabular}{l|c|c}
\hline
Metric & SINGAPO & Ours \\
\hline
Occupancy\_recall $\uparrow$ & / & \textbf{98.94\%} \\
Bbox center $\downarrow$ & 0.0440 & \textbf{0.0357} \\
Bbox size $\downarrow$ & \textbf{0.0651} & 0.0832 \\
Part type $\uparrow$ & \textbf{97.89\%} & 96.27\% \\
Joint type $\uparrow$ & 97.37\% & \textbf{99.17\%} \\
Joint axis $\downarrow$ & 1.29$^\circ$ & \textbf{1.14$^\circ$} \\
Joint origin $\downarrow$ & 0.39 & \textbf{0.10} \\
Joint range (angle) $\downarrow$ & \textbf{6.73$^\circ$} & 7.37$^\circ$ \\
Joint range (translation) $\downarrow$ & 0.0265 & \textbf{0.0159} \\
\hline
\end{tabular}
\label{tab:vae_comparison}
\end{table}

\begin{table}[ht]
\caption{Standard deviation ($std$) of predicted articulation parameters across voxels within each part.}
\centering
\begin{tabular}{l|c}
\hline
Metric & $std$ \\
\hline
Bbox center $\downarrow$ & {[0.0211, 0.0159, 0.0277]} \\
Bbox size $\downarrow$ & {[0.0377, 0.0260, 0.0391]} \\
Joint axis $\downarrow$ & {[0.1012, 0.0972, 0.0875]} \\
Joint origin $\downarrow$ & {[0.0392, 0.0233, 0.0384]} \\
Joint range (angle) $\downarrow$ & {[6.96$^\circ$, 7.56$^\circ$]} \\
Joint range (translation) $\downarrow$ & {[0.007, 0.0123]} \\
\hline
\end{tabular}
\label{tab:std}
\end{table}

\paragraph{Generalization to unseen dataset}
To evaluate the generalization capability of our method, we test on the ACD dataset, which contains articulated objects with part configurations and motion patterns not seen during training. As shown in Table~\ref{tab:quant_pm_acd} and the last two rows in Fig.~\ref{fig:visual_compasion}, our method significantly outperforms all baselines in both quantitative and qualitative comparisons.

Besides, to evaluate our method's applicability in real-world scenarios, we conduct qualitative experiments using real-captured images of articulated household objects. As shown in Fig.~\ref{fig:teaser}, our method successfully synthesizes plausible voxel structures and decodes them into textured 3D Gaussians that exhibit coherent geometry and physically realistic articulation.
Despite being trained on synthetic datasets, our model generalizes well to real images, capturing fine-grained part semantics and motion behaviors. 

\paragraph{Effectiveness of articulation-aware fine-tuning}
To assess the impact of our articulation-aware fine-tuning strategy, we conduct an ablation study by comparing models trained with and without this component, as shown in Table~\ref{tab:quant_abla} and Fig.~\ref{fig:stage2_results}. Removing fine-tuning results in noticeable performance degradation, particularly in AS-$d_{\text{CD}}$ and FID, where texture artifacts and inconsistencies in articulated regions become prominent. In contrast, applying articulation-aware supervision leads to lower geometric error and improved perceptual realism. These findings highlight the importance of adapting the appearance decoder to articulation-dependent visibility changes. 

\paragraph{Inference efficiency.}
We evaluate the runtime performance of our method on an NVIDIA A6000 GPU. The total inference time for generating an articulated 3D object consists of three stages: sampling the articulation-aware voxel structure (16.25 seconds), sampling voxel-level appearance features (9.54 seconds), and decoding the final 3D Gaussian splats (0.06 seconds), resulting in an overall runtime of approximately 25.85 seconds per object.
In comparison, the baseline method SINGAPO requires around 2.9 seconds per object under the same hardware setting.

\paragraph{Failure cases and limitations}
Despite the promising results, our method still has several limitations. First, we evaluate our framework on articulated objects with relatively simple kinematic structures, such as the furniture categories from the PartNet-Mobility dataset and ACD datasets. 
While some real-scene results in Fig.~\ref{fig:teaser} and quantitative results on ACD dataset demonstrate a certain degree of generalization to unseen, yet similar object categories within the training domains. We acknowledge that our model’s generalization could be further improved with a larger dataset.
Second, although our framework models the object holistically and preserves global part coherence, it relies on accurate part-level bounding boxes and voxel-level semantic labels to segment individual parts. In cases where these annotations are imprecise or inconsistently sampled, part segmentation quality degrades, which may result in distorted part geometry or incorrect motion behavior (see Fig.~\ref{fig:limitations}).

\label{sec:experiments}

\section{Conclusion}
We presented \nickname, a unified generative framework for synthesizing human-made articulated 3D objects with fine-grained geometry, motion semantics, and realistic appearance. By embedding articulation-aware voxel structures into a compact latent space and leveraging structured diffusion priors, our method supports controllable generation conditioned on a single image. To address articulation-aware visibility changes, we introduced a fine-tuning strategy that significantly improves appearance fidelity in both external and internal regions.
Extensive experiments on standard benchmarks demonstrate that ArtiLatent achieves state-of-the-art performance in geometric accuracy, motion plausibility, and visual realism. Our approach opens new possibilities for scalable articulated 3D content creation, interactive editing, and robotic simulation. Future work includes building larger and more diverse datasets, exploring generalization to more natural dynamics, scaling to large object libraries, and integrating physical constraints for simulation-ready assets.

\label{sec:conclusions}

% Bibliography
\bibliographystyle{ACM-Reference-Format}
\bibliography{sample-bibliography}

% Appendix
\appendix

\begin{figure*}[t]
    \centering
    \includegraphics[width=\linewidth]{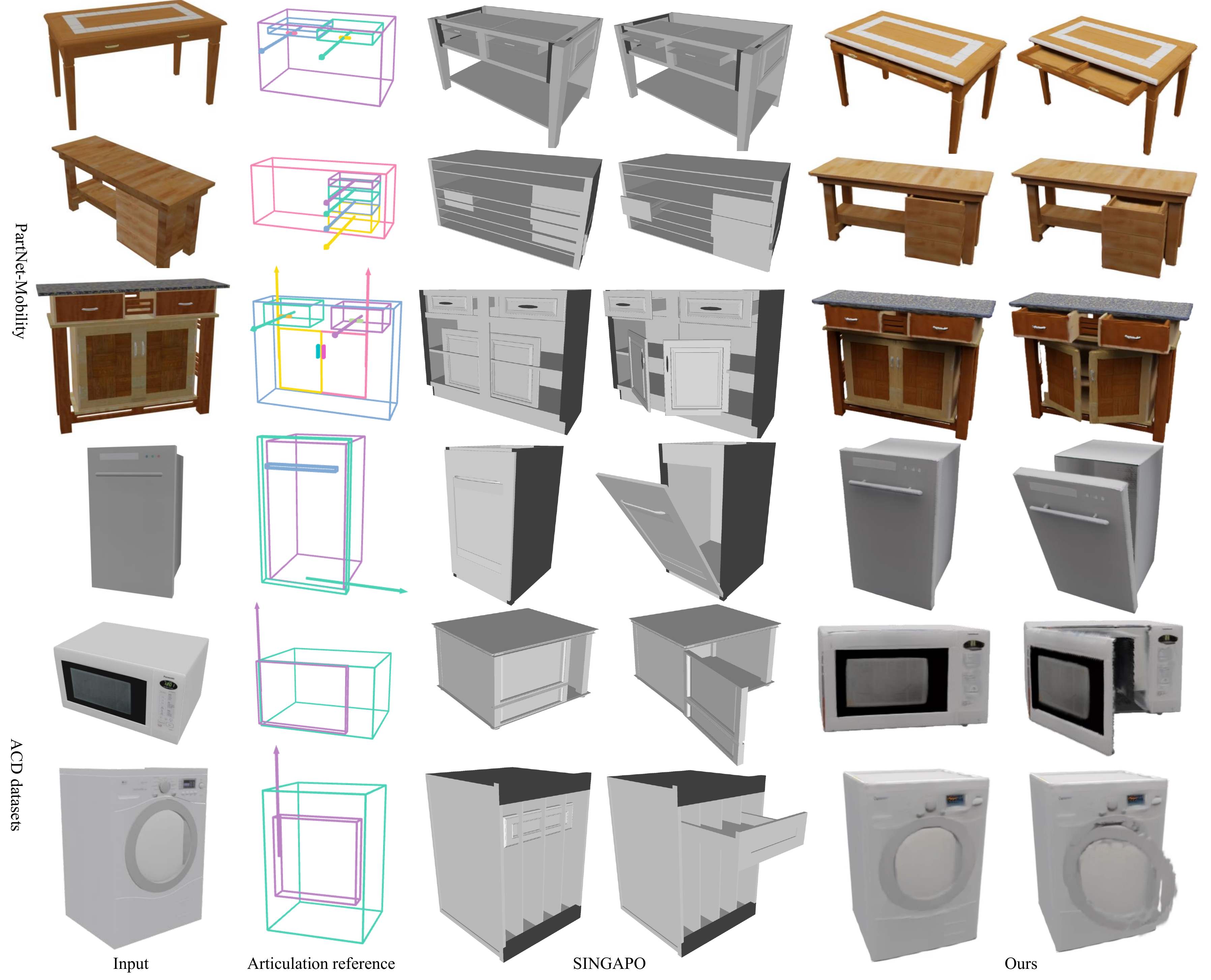}
    \caption{Qualitative comparison across different categories from the PartNet-Mobility and ACD datasets. The first column shows the input image, and the second column visualizes the ground-truth abstract articulation as a reference. Each object is displayed in both its resting and articulated states.}
    \label{fig:visual_compasion}
\end{figure*}

\begin{figure*}[t]
  \centering
  \includegraphics[width=\linewidth]{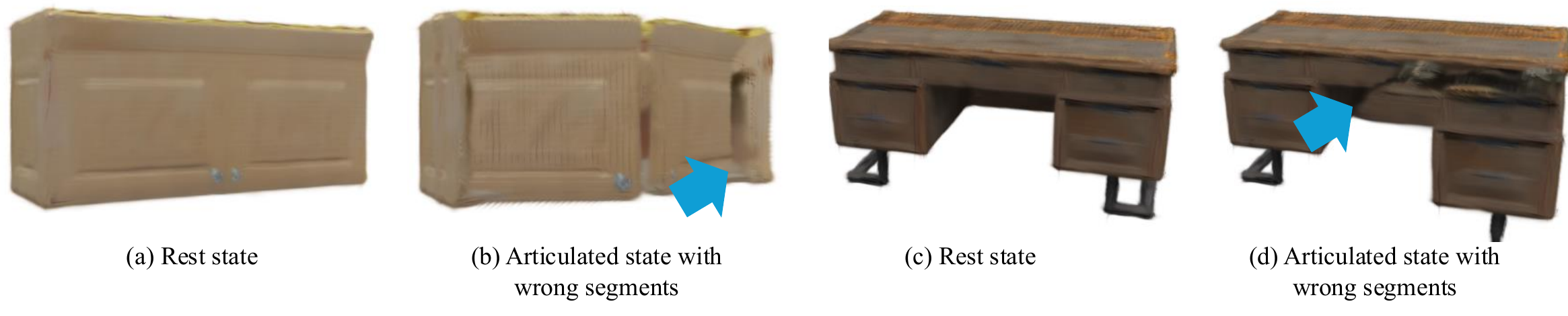}
    \caption{
    Failure cases due to incorrect part segmentation.
    We show two examples where inaccurate voxel-part segmentation leads to unrealistic articulation. In both cases, the generated objects in the rest state (a, c) appear structurally correct, but in the articulated state (b, d), incorrect part grouping results in implausible deformations and motion artifacts (highlighted with blue arrows). 
    }
  \label{fig:limitations}
\end{figure*}

\begin{figure*}[t]
  \centering
  \includegraphics[width=0.78\linewidth]{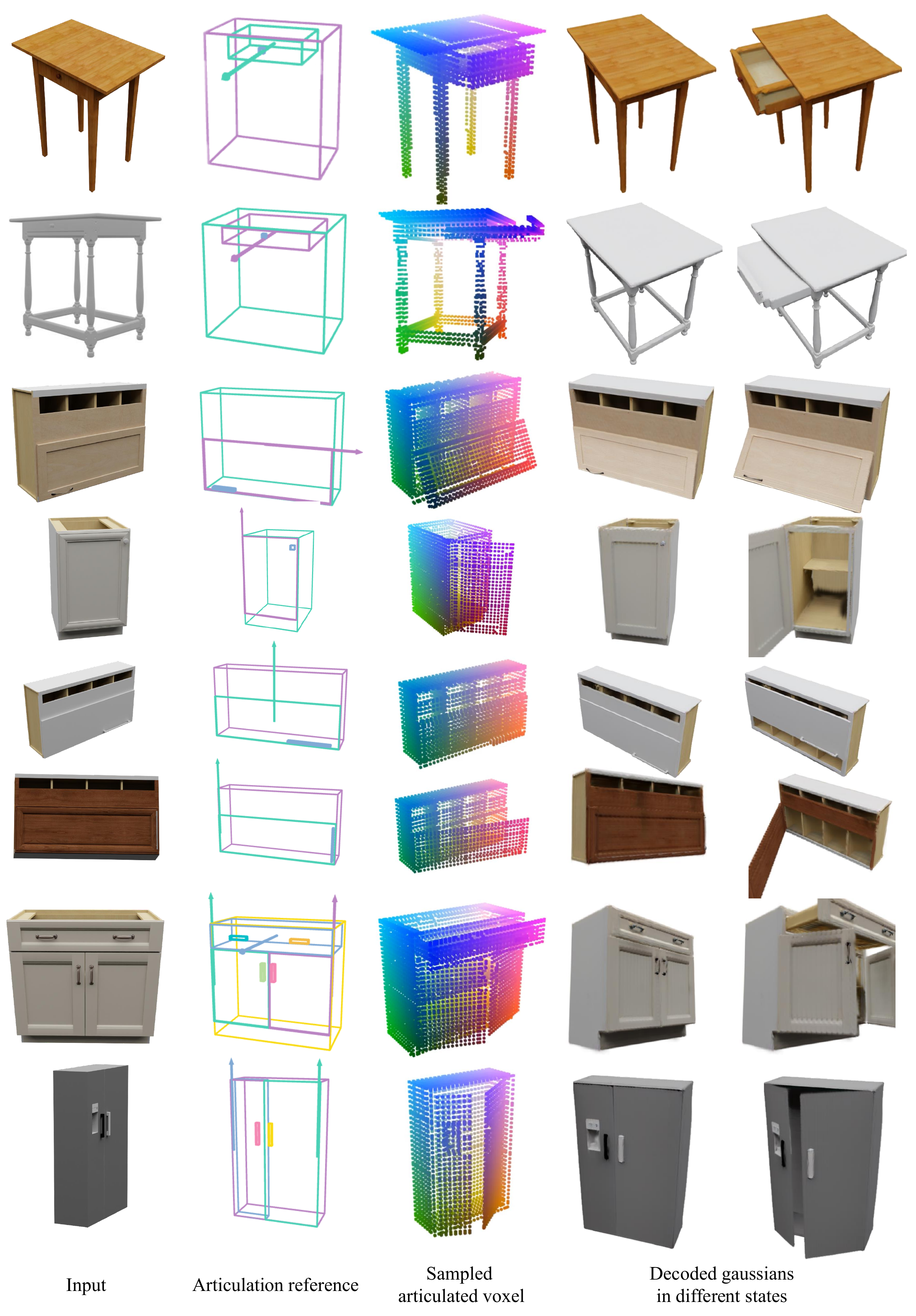}
    \caption{
Additional visual results on articulated 3D object generation.
For each example, we show the input condition (1-st column), the ground-truth articulation reference (2-nd column), the sampled articulation-aware voxel representation (3-th column), and the decoded 3D Gaussian splats in different articulation states (4-th and 5-th columns). Our method consistently produces coherent geometry, realistic part appearance, and physically plausible articulation across diverse object types.
}
  \label{fig:additional_results}
\end{figure*}

% \section{Switching Times}

% In this appendix, we measure the channel switching time of Micaz
% \cite{CROSSBOW} sensor devices.  In our experiments, one mote
% alternatingly switches between Channels~11 and~12. Every time after
% the node switches to a channel, it sends out a packet immediately and
% then changes to a new channel as soon as the transmission is finished.
% We measure the number of packets the test mote can send in 10 seconds,
% denoted as $N_{1}$. In contrast, we also measure the same value of the
% test mote without switching channels, denoted as $N_{2}$. We calculate
% the channel-switching time $s$ as
% \begin{displaymath}%
% s=\frac{10}{N_{1}}-\frac{10}{N_{2}}.
% \end{displaymath}%
% By repeating the experiments 100 times, we get the average
% channel-switching time of Micaz motes: 24.3\,$\mu$s.

\end{document}